\documentclass[10pt,twocolumn,letterpaper]{article}

\usepackage[pagenumbers]{cvpr} 

\usepackage{graphicx}
\usepackage{amsmath}
\usepackage{amssymb}
\usepackage{booktabs}
\usepackage{url}

\usepackage{utfsym}
\usepackage{pifont}
\usepackage{soul} 
\usepackage{dsfont}

\usepackage[pagebackref,breaklinks,colorlinks]{hyperref}

\usepackage{multirow}
\usepackage{pifont}
\usepackage{colortbl}
\usepackage[capitalize,noabbrev]{cleveref}
\usepackage{mathrsfs}
\crefname{section}{Sec.}{Secs.}
\Crefname{section}{Section}{Sections}
\Crefname{table}{Table}{Tables}
\crefname{table}{Tab.}{Tabs.}

\makeatletter
\newcommand{\thickhline}{%
	\noalign {\ifnum 0=`}\fi \hrule height 1pt
	\futurelet \reserved@a \@xhline
}
\makeatother

\def\cvprPaperID{*****} 
\def\confName{CVPR}
\def\confYear{2024}

\begin{document}

\title{2nd Place Solution for MeViS Track in CVPR 2024 PVUW Workshop: Motion Expression guided Video Segmentation}  

\author{Bin Cao$^{1,2,3}$\thanks{Equal contribution.}{\qquad} Yisi Zhang$^{4}${\qquad} Xuanxu Lin$^{2}${\qquad}Xingjian He$^{1}${\qquad}Bo Zhao$^{3}${\qquad}Jing Liu$^{1,2}$\thanks{ Corresponding author.}
\vspace{3mm}\\
$^{1}$Institute of Automation, Chinese Academy of Sciences (CASIA)\qquad \\
$^{2}$School of Artificial Intelligence, University of Chinese Academy of Sciences (UCAS) \\
$^{3}$Beijing Academy of Artificial Intelligence (BAAI) \\
$^{4}$University of Science and Technology Beijing (USTB) \\
Team: CASIA\_IVA
}

\maketitle
\begin{abstract}
Motion Expression guided Video Segmentation is a challenging task that aims at segmenting objects in the video based on natural language expressions with motion descriptions. Unlike the previous referring video object segmentation (RVOS), this task focuses more on the motion in video content for language-guided video object segmentation, requiring an enhanced ability to model longer temporal, motion-oriented vision-language data. In this report, based on the RVOS methods, we successfully introduce mask information obtained from the video instance segmentation model as preliminary information for temporal enhancement and employ SAM for spatial refinement. Finally, our method achieved a score of 49.92 $\mathcal{J}\&\mathcal{F}$ in the validation phase and 54.20 $\mathcal{J}\&\mathcal{F}$ in the test phase, securing the final ranking of 2nd in the MeViS Track at the CVPR 2024 PVUW Challenge.

\end{abstract}

\section{Introduction}
Referring Video Object Segmentation (ROVS) aims to segment and track the target object referred by the given language description.
This emerging task has attracted significant attention due to its potential applications in video editing and human-agent interaction.

Motion Expression-guided Video Segmentation is a challenging task based on the RVOS task. Given videos and motion-oriented language expressions obtained from the dataset MeViS \cite{ding2023mevis}, an embodied agent is needed to segment the corresponding described one target or multiple objects in the video. Compared to the conventional RVOS datasets like Ref-Youtube-VOS \cite{seo2020urvos} and Ref-DAVIS17 \cite{khoreva2019video}, MeViS presents more complex expressions that include motion information rather than merely simple spatial location descriptions. Consequently, MeViS necessitates that the agent comprehends both temporal and spatial information within video clips to effectively correlate with motion expressions. Furthermore, MeViS extends the referring task to include language expressions that match multiple targets, making MeViS more challenging and reflective of real-world scenarios. 

With the development of deep learning, there are studies dealing with the RVOS task. For example, some studies \cite{seo2020urvos,bellver2023closer} try to deal with the task from a per-frame perspective, they transfer the referring image segmentation methods \cite{huang2020referring, luo2020multi, ding2021vision} into video domain, whether through a per-frame mask propagation manner or based on history memory attention to predict the mask of the current frame. Most recent methods, e.g. \cite{wu2022language,miao2023spectrum,luo2024soc,yan2024referred,wu2023onlinerefer,botach2022end}, focus on a unified framework that employs language as queries to segment and track the referred object simultaneously. By effectively building correlations between expressions and multi-frame visual features, they achieve promising results across multiple RVOS benchmarks. Some recent methods, e.g.\cite{yan2023universal,wu2023general,li2024univs}, unify various kinds of object-level tasks, such as MOT, VIS, RVOS and RES, into a single framework to present an object-centric foundation model. However, these methods still encounter the issue of inconsistent predicted results across multiple frames. While some recent studies on Video Instance Segmentation (VIS) task\cite{yang2019video}, which emphasizes segmenting different instances in the given video, have shown promising results in dealing with prediction inconsistent problem. Furthermore, the emergence of SAM \cite{kirillov2023segment} also provides strong segmentation tools for refinement. 

\begin{figure*}[t]
  \centering
   \includegraphics[width=1.0\linewidth]{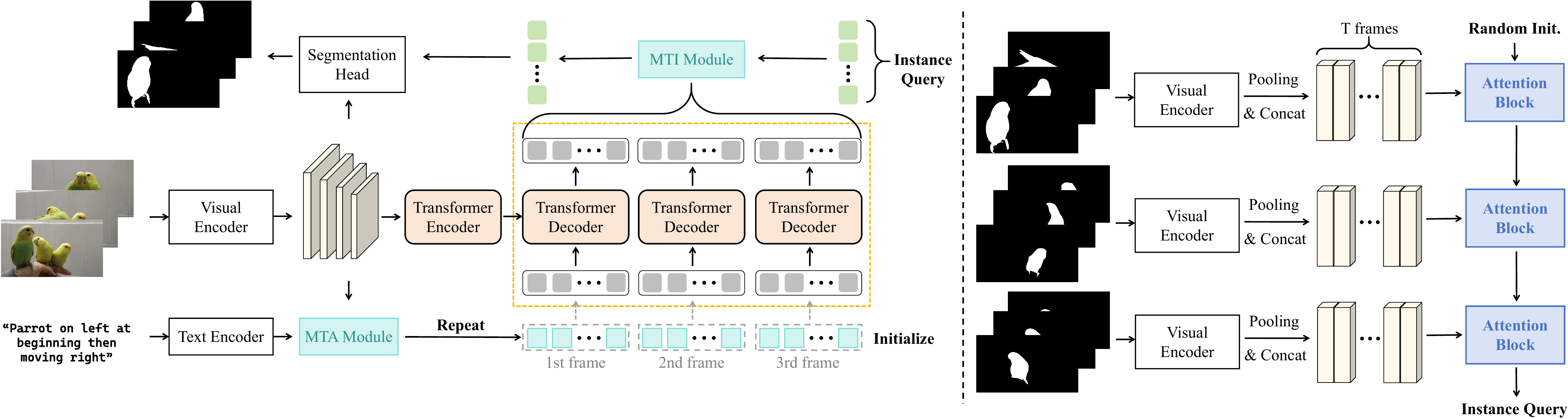}
   \caption{\textbf{The overall architecture of our solution.} We employ MUTR as our basic model (\textbf{Left}), which contains a visual/text backbone, transformer-based encoder and decoder, MTI module, and MTA module. We attempt to introduce instance masks to improve the consistency of prediction results. We employ an attention block and a sequential mechanism to aggregate instance information into a query (\textbf{Right}).}
   \label{fig:model_architecture}
\end{figure*}

Thanks to the superior performance of DVIS \cite{zhang2023dvis} from the VIS task, MUTR \cite{yan2024referred} from the RVOS task, and HQ-SAM \cite{ke2024segment}, our method achieves a score of 49.92 $\mathcal{J}\&\mathcal{F}$ in the validation phase and 54.20 $\mathcal{J}\&\mathcal{F}$ in the test phase, securing the final ranking of 2nd in the MeViS Track at the CVPR 2024 PVUW Challenge.
\label{sec:intro}

\section{Method}
\subsection{Overview}
The architecture of our model is shown in \cref{fig:model_architecture}. We employ a DETR-based model as our basic architecture. To improve the consistency of predicted results, we attempt to introduce proposal instance masks into the model for query initialization. After prediction, we employ HQ-SAM to refine prediction masks by sampling key points as prompts. 

\subsection{Basic Model Architecture}
MUTR (\textbf{M}ultimodal \textbf{U}nified \textbf{T}emporal transformer for \textbf{R}eferring video object segmentation) was proposed in \cite{yan2024referred} and has shown superior performance on Ref-Youtube-VOS and Ref-DAVIS17. MUTR adopts a DETR-like style model. Compared with other methods, MUTR introduces two core modules, i.e. MTI (\textbf{M}ulti-object \textbf{T}emporal \textbf{I}nteraction module), MTA (\textbf{M}ulti-scale \textbf{T}emporal \textbf{A}ggregation module). 

To conduct multi-modal interaction and fusion, the MTA module consists of sequential cross-attention blocks, which takes text feature and multi-scale visual features as input and progressively captures temporal visual information. After that, the class token of output multi-modal tokens is repeated to initialize object queries of transformer decoder. Considering that transformer encoder and decoder process video in a frame-independent manner without temporal modeling, MTI is designed to perform object-wise interaction. MTI module contains an encoder and a decoder. MTI encoder conducts temporal interaction of the same object across frames and MTI decoder is designed to aggregate information of object queries. MTI encoder takes the output of transformer decoder as input and performs self-attention for the same object across multiple frames to conduct temporal interaction. MTI decoder consists of a cross-attention layer, a self-attention layer, and a FFN layer. In MTI decoder, a set of video-wise query $\mathcal{Q}$ with random initialization is adopted to associate objects. MTI decoder takes the output of MTI encoder as key and value and conducts cross-attention with video-wise query.

\begin{table*}[ht]
\centering
\caption{\textbf{Ablation Experiment Results on MeViS Validation Set} We ablate multiple design choices, including sampling method, HQ-SAM, and introducing instance masks for query initialization. In all experiments, we adopt the pre-trained weights of MUTR as initialization and fine-tune model on MeViS.}
\label{tab:ablation}
\vspace{5pt}
\resizebox{0.9\linewidth}{!}{
\begin{tabular}{c|c|c|c|c|ccc}
\toprule[1.5pt]
Method  &Backbone & Sampling Method  & Instance Masks  & HQ-SAM    & $\mathcal{J}\&\mathcal{F}$   & $\mathcal{J}$   & $\mathcal{F}$      \\ 
\midrule
\multirow{6}{*}{MUTR}   & \multirow{6}{*}{Swin-L}                           
& \multirow{2}{*}{Local Sampling}  &\ding{55}  &\ding{55}  &48.40   &44.87  &51.94 
\\
& &  &\ding{55} & {\checkmark} & \cellcolor{gray!25}48.66  & \cellcolor{gray!25}45.86 & \cellcolor{gray!25}51.46    \\ 
\cmidrule(lr){3-8}
& & \multirow{2}{*}{Global Sampling}  &\ding{55}  &\ding{55}  &49.11   &45.89  &52.33 
\\
& &  &\ding{55} & {\checkmark} & \cellcolor{gray!25}49.50  & \cellcolor{gray!25}46.91 & \cellcolor{gray!25}52.09    \\ 
\cmidrule(lr){3-8}
& & \multirow{2}{*}{Global Sampling}  & {\checkmark}  &\ding{55}  &49.62   &46.38  &52.85 
\\
& & & {\checkmark} & {\checkmark} & \cellcolor{gray!25}\textbf{49.92}  & \cellcolor{gray!25}\textbf{47.30} & \cellcolor{gray!25}\textbf{52.54}    \\ 
        
\bottomrule[1.5pt]

\end{tabular}
}
\end{table*}

\subsection{Instance Masks for Query Initialization}
While MUTR achieves superior performance on RVOS, prediction results from MUTR still suffer from inconsistency and incompleteness. Meantime, some recent studies on VIS show promising results to solve this issue. Therefore, we attempt to introduce instance mask information into a DETR-based model to improve the consistency and completeness of prediction results.

Specifically, we attempt to introduce instance masks to initialize the video-wise query $\mathcal{Q}$ in MTI decoder. Thanks to the superior performance of DVIS on VIS, we employ DVIS for mask generation, which extracts all instance masks in a video clip as follows:
\begin{equation}
    m_{i} = \text{DVIS}(\text{I}), \ m_{i} \in \mathbb{R}^{\ T \times H \times W}
\end{equation}
where $\text{I} \in \mathbb{R}^{\ T \times H \times W \times \text{3}}$ is the input video clip, $m = \{m_{i}\}^{K}_{i=1}$ denotes the set of instance masks, $K$ is the number of instances in a video clip and $T$ is the number of frames of a video clip.

Next, we utilize a visual encoder to extract multi-scale visual features of instance masks.
\begin{equation}
    \mathcal{F}_{i,j} = \text{Visual\_Backbone}(m_{i}), \ \mathcal{F}_{i,j} \in \mathbb{R}^{\ T \times h_{j} \times w_{j} \times c_{j} }
\end{equation}
where $c_{j}$ is the channel of $j$ level visual feature.
After feature extraction, we utilize a linear projection layer on multi-scale visual features to align dimension with video features and perform average pooling along spatial dimension to obtain instance features as follows: 
\begin{equation}
    \mathcal{F}^{'}_{i,j} = \text{Pooling}(\text{Proj}(\mathcal{F}_{i,j})), \  \mathcal{F}_{i,j} \in \mathbb{R}^{\ T \times C }
\end{equation}
where $C$ is the channel of video feature.

For simplicity, we only explain our solution utilizing the single-level visual feature. To aggregate all instance information into an instance query, we design an attention block and adapt sequential mechanisms as follows: 
\begin{equation}
    \mathcal{Q}_{i} = \text{Block}(\mathcal{Q}_{i-1},\mathcal{F}^{'}_{i}), \ 1 \leq i \leq K 
\end{equation}
where ${Q}_{i} \in \mathbb{R}^{\ N \times C}$ is the instance query and $N$ is the number of queries. ${Q}_{0}$ is randomly initialized. The designed attention block consists of a cross-attention layer, a set of self-attention layers, and FFN layers. After that, we utilize this query with instance information to replace the randomly initialized video-wise query fed to MTI decoder.

\subsection{HQ-SAM for Spatial Refinement}
Since SAM has shown its great ability in segmenting objects, it could serve as a spatial refiner for better results. Specifically, in this report, we adopt HQ-SAM \cite{ke2024segment} with ViT-L as our mask refiner. Given the predicted result from MUTR of each clip, we first determine the coordinates of the bounding box by selecting the maximum and minimum horizontal and vertical coordinates of the points along the boundary of the mask. Next, we uniformly sample 10 coordinates within the predicted mask as positive points and 5 coordinates out of the mask but within the bounding box as negative points. The sampled points are then fed into the mask decoder of HQ-SAM as prompts to generate the refined masks.

\section{Experiments}
\subsection{Dataset and Metrics}
\noindent{\textbf{Datasets.}} We fine-tune and evaluate our solution on MeViS, a large-scale dataset for motion guided video segmentation. It contains 2,006 videos with 28,570 language expressions in total. These videos are divided into 1,662 videos for training, 50 videos for offline evaluation, 140 videos for online evaluation, and other videos for competition.

\noindent{\textbf{Metrics.}} We adopt standard evaluation metrics for MeViS: region similarity ($\mathcal{J}$), contour accuracy ($\mathcal{F}$), and their average value ($\mathcal{J}\&\mathcal{F}$).

\subsection{Sampling Method}
In the training phase, previous work in RVOS sample frames around a center point, we named this method local sampling. This method only allows model to access part of the video. However, motion expression guided video segmentation requires a video-level representation. Therefore, we attempt to divide the entire video into a set of segments and sample one frame randomly in each segment, and aggregate sampling frames to obtain a video clip fed into model. We refer this approach as global sampling, which enables model to access frames across the entire video.


\subsection{Implement Details}
We adopt the pre-trained weights of MUTR as initialization and fine-tune model on MeViS. The number of sampling frames is 5. The model is optimized by AdamW optimizer. The batch size is 1 and the accumulation step is 2. In post-process, we employ HQ-SAM with VIT-L backbone utilizing default parameters in \cite{ke2024segment}. 

\subsection{Ablation Experiments}
To validate the effectiveness of introducing instance mask for query initialization, sampling method and HQ-SAM, we conduct simple ablation experiments. We adopt the pre-trained weight of MUTR for weight initialization and fine-tuning model on MeViS. Experiment results are shown in \cref{tab:ablation}. It is noted that utilizing HQ-SAM for refinement brings an improvement on $\mathcal{J}$ while a drop about $\mathcal{F}$. However, utilizing HQ-SAM for refinement still brings an improvement on $\mathcal{J}\&\mathcal{F}$. Compared with the previous sampling method, the proposed sampling method brings a significant improvement about 0.69 $\mathcal{J}\&\mathcal{F}$. After introducing instance masks for query initialization, the performance improved from 49.11 $\mathcal{J}\&\mathcal{F}$ to 49.62 $\mathcal{J}\&\mathcal{F}$. Finally, when we combine all of the above methods, model achieves the best performance 49.92 $\mathcal{J}\&\mathcal{F}$.

\subsection{Competition Results}
Finally, we submit our best solution and achieve 54.20 $\mathcal{J}\&\mathcal{F}$ ( 50.97 $\mathcal{J}$ and 57.43 $\mathcal{F}$ ) on test phase, which ranks the 2nd place for MeViS Track in CVPR 2024 PVUW.

{\small
\bibliographystyle{ieee_fullname}
\bibliography{egbib}

\begin{thebibliography}{10}\itemsep=-1pt

\bibitem{bellver2023closer}
Miriam Bellver, Carles Ventura, Carina Silberer, Ioannis Kazakos, Jordi Torres, and Xavier Giro-i Nieto.
\newblock A closer look at referring expressions for video object segmentation.
\newblock {\em Multimedia Tools and Applications}, 82(3):4419--4438, 2023.

\bibitem{botach2022end}
Adam Botach, Evgenii Zheltonozhskii, and Chaim Baskin.
\newblock End-to-end referring video object segmentation with multimodal transformers.
\newblock In {\em Proceedings of the IEEE/CVF Conference on Computer Vision and Pattern Recognition}, pages 4985--4995, 2022.

\bibitem{ding2023mevis}
Henghui Ding, Chang Liu, Shuting He, Xudong Jiang, and Chen~Change Loy.
\newblock Mevis: A large-scale benchmark for video segmentation with motion expressions.
\newblock In {\em Proceedings of the IEEE/CVF International Conference on Computer Vision}, pages 2694--2703, 2023.

\bibitem{ding2021vision}
Henghui Ding, Chang Liu, Suchen Wang, and Xudong Jiang.
\newblock Vision-language transformer and query generation for referring segmentation.
\newblock In {\em Proceedings of the IEEE/CVF International Conference on Computer Vision}, pages 16321--16330, 2021.

\bibitem{huang2020referring}
Shaofei Huang, Tianrui Hui, Si Liu, Guanbin Li, Yunchao Wei, Jizhong Han, Luoqi Liu, and Bo Li.
\newblock Referring image segmentation via cross-modal progressive comprehension.
\newblock In {\em Proceedings of the IEEE/CVF conference on computer vision and pattern recognition}, pages 10488--10497, 2020.

\bibitem{ke2024segment}
Lei Ke, Mingqiao Ye, Martin Danelljan, Yu-Wing Tai, Chi-Keung Tang, Fisher Yu, et~al.
\newblock Segment anything in high quality.
\newblock {\em Advances in Neural Information Processing Systems}, 36, 2024.

\bibitem{khoreva2019video}
Anna Khoreva, Anna Rohrbach, and Bernt Schiele.
\newblock Video object segmentation with language referring expressions.
\newblock In {\em Computer Vision--ACCV 2018: 14th Asian Conference on Computer Vision, Perth, Australia, December 2--6, 2018, Revised Selected Papers, Part IV 14}, pages 123--141. Springer, 2019.

\bibitem{kirillov2023segment}
Alexander Kirillov, Eric Mintun, Nikhila Ravi, Hanzi Mao, Chloe Rolland, Laura Gustafson, Tete Xiao, Spencer Whitehead, Alexander~C Berg, Wan-Yen Lo, et~al.
\newblock Segment anything.
\newblock In {\em Proceedings of the IEEE/CVF International Conference on Computer Vision}, pages 4015--4026, 2023.

\bibitem{li2024univs}
Minghan Li, Shuai Li, Xindong Zhang, and Lei Zhang.
\newblock Univs: Unified and universal video segmentation with prompts as queries.
\newblock {\em arXiv preprint arXiv:2402.18115}, 2024.

\bibitem{luo2020multi}
Gen Luo, Yiyi Zhou, Xiaoshuai Sun, Liujuan Cao, Chenglin Wu, Cheng Deng, and Rongrong Ji.
\newblock Multi-task collaborative network for joint referring expression comprehension and segmentation.
\newblock In {\em Proceedings of the IEEE/CVF Conference on computer vision and pattern recognition}, pages 10034--10043, 2020.

\bibitem{luo2024soc}
Zhuoyan Luo, Yicheng Xiao, Yong Liu, Shuyan Li, Yitong Wang, Yansong Tang, Xiu Li, and Yujiu Yang.
\newblock Soc: Semantic-assisted object cluster for referring video object segmentation.
\newblock {\em Advances in Neural Information Processing Systems}, 36, 2024.

\bibitem{miao2023spectrum}
Bo Miao, Mohammed Bennamoun, Yongsheng Gao, and Ajmal Mian.
\newblock Spectrum-guided multi-granularity referring video object segmentation.
\newblock In {\em Proceedings of the IEEE/CVF International Conference on Computer Vision}, pages 920--930, 2023.

\bibitem{seo2020urvos}
Seonguk Seo, Joon-Young Lee, and Bohyung Han.
\newblock Urvos: Unified referring video object segmentation network with a large-scale benchmark.
\newblock In {\em Computer Vision--ECCV 2020: 16th European Conference, Glasgow, UK, August 23--28, 2020, Proceedings, Part XV 16}, pages 208--223. Springer, 2020.

\bibitem{wu2023onlinerefer}
Dongming Wu, Tiancai Wang, Yuang Zhang, Xiangyu Zhang, and Jianbing Shen.
\newblock Onlinerefer: A simple online baseline for referring video object segmentation.
\newblock In {\em Proceedings of the IEEE/CVF International Conference on Computer Vision}, pages 2761--2770, 2023.

\bibitem{wu2023general}
Junfeng Wu, Yi Jiang, Qihao Liu, Zehuan Yuan, Xiang Bai, and Song Bai.
\newblock General object foundation model for images and videos at scale.
\newblock {\em arXiv preprint arXiv:2312.09158}, 2023.

\bibitem{wu2022language}
Jiannan Wu, Yi Jiang, Peize Sun, Zehuan Yuan, and Ping Luo.
\newblock Language as queries for referring video object segmentation.
\newblock In {\em Proceedings of the IEEE/CVF Conference on Computer Vision and Pattern Recognition}, pages 4974--4984, 2022.

\bibitem{yan2023universal}
Bin Yan, Yi Jiang, Jiannan Wu, Dong Wang, Ping Luo, Zehuan Yuan, and Huchuan Lu.
\newblock Universal instance perception as object discovery and retrieval.
\newblock In {\em Proceedings of the IEEE/CVF Conference on Computer Vision and Pattern Recognition}, pages 15325--15336, 2023.

\bibitem{yan2024referred}
Shilin Yan, Renrui Zhang, Ziyu Guo, Wenchao Chen, Wei Zhang, Hongyang Li, Yu Qiao, Hao Dong, Zhongjiang He, and Peng Gao.
\newblock Referred by multi-modality: A unified temporal transformer for video object segmentation.
\newblock In {\em Proceedings of the AAAI Conference on Artificial Intelligence}, volume~38, pages 6449--6457, 2024.

\bibitem{yang2019video}
Linjie Yang, Yuchen Fan, and Ning Xu.
\newblock Video instance segmentation.
\newblock In {\em Proceedings of the IEEE/CVF international conference on computer vision}, pages 5188--5197, 2019.

\bibitem{zhang2023dvis}
Tao Zhang, Xingye Tian, Yu Wu, Shunping Ji, Xuebo Wang, Yuan Zhang, and Pengfei Wan.
\newblock Dvis: Decoupled video instance segmentation framework.
\newblock In {\em Proceedings of the IEEE/CVF International Conference on Computer Vision}, pages 1282--1291, 2023.

\end{thebibliography}
}

\end{document}